# Smoothness and Structure Learning by Proxy

**Benjamin Yackley**  BENJ@CS.UNM.EDU
Department of Computer Science, University of New Mexico, Albuquerque, NM 87131

**Terran Lane**  TERRAN@CS.UNM.EDU
Department of Computer Science, University of New Mexico, Albuquerque, NM 87131

## Abstract

As data sets grow in size, the ability of learning methods to find structure in them is increasingly hampered by the time needed to search the large spaces of possibilities and generate a score for each that takes all of the observed data into account. For instance, Bayesian networks, the model chosen in this paper, have a super-exponentially large search space for a fixed number of variables. One possible method to alleviate this problem is to use a proxy, such as a Gaussian Process regressor, in place of the true scoring function, training it on a selection of sampled networks. We prove here that the use of such a proxy is well-founded, as we can bound the smoothness of a commonly-used scoring function for Bayesian network structure learning. We show here that, compared to an identical search strategy using the network's exact scores, our proxy-based search is able to get equivalent or better scores on a number of data sets in a fraction of the time.

## 1. Introduction

Probabilistic graphical models such as Bayesian networks (Koller & Friedman, 2009), which explain patterns in data through dependence relations among variables, are a useful tool because of the visibility and ease of interpretation of the models and because of the ability to estimate distributions given known values. However, generating these models from observed data runs into problems in a number of ways. If the data set has too many variables, the number of possible models grows exponentially, and if there are too many data points, it becomes more time-intensive to analyze the relationships between them. Although this growth is only linear in the number of data points, modern data sets run into the gigabytes or larger. What is needed is a way to separate the size of the data set from the search process, and this is the purpose of the proxy. By training a function approximator on exact scores of a random sample of networks, we can then use that proxy in the search without ever needing to go back to the original data.

We prove here that the BDe scoring function is reasonably smooth over a properly chosen topology, and this fact motivates the use of a Gaussian process regressor as a proxy to the exact function. Once the proxy is built, the original data set need never be touched again, and the search itself can proceed extremely quickly. As our results show, even taking into account the additional time needed to score the training samples and generate the proxy from them, we are often able to generate better-scoring models than an exact-scoring search in a smaller amount of time.

## 2. Background

### 2.1. Bayesian networks

A Bayesian network (Heckerman et al., 1995) is a statistical model used to represent probabilistic relationships among a set of variables as a directed acyclic graph, where the distribution of a single variable is defined in terms of the values of that variable's parents in the graph. Bayesian networks are commonly used to infer distributions over unobserved or query variables given known values for others (for instance, spam classification (Sebastiani & Ramoni, 2001) or disease diagnosis (Burge et al., 2009)). The process of learning a Bayesian network given a set of observed data is difficult, and is in fact NP-complete in the case of finding an exact optimum (Chickering, 1996; Chickering et al., 1995); most techniques are still limited in practice in the number of variables they can handle at once. One key component of many of these algorithms is a score function which, given a fixed data set, maps individual graphs onto real numbers; the optimal network is the one whose graph has the highest score. In other words, the function $sc(G|D)$ maps the space of directed acyclic graphs $\mathbb{G}_n$ on $n$ nodes, one for each variable $x_1 \ldots x_n$, along with a data set $D \in \mathbb{N}^{m \times n}$ with $m$ i.i.d. observations of those variables, and the desired output of the search process is $\arg\max_G sc(G|D)$.

### 2.2. The BDe score

There are many Bayesian network scoring functions one could use as a basis for a search, but the BDe score (Heckerman et al., 1995), has several desired properties. First, it is decomposable, meaning that it can be expressed as a function of independent components, one for each node in the graph. Second, it is a Bayesian formulation that allows us to enforce a prior belief over graph structures independent of the data itself. Finally, the structure of the BDe score is straightforward, requiring only counts of queries over the data (which can be made easier using an ADTree (Anderson & Moore, 1998), as described below) and the log-Gamma function, which itself is easily approximated numerically. The form of the BDe score is:

$$sc(G|D) = \prod_{i=1}^{n} \prod_{j \in C(x_i)} \frac{\Gamma(\lambda_{ij})}{\Gamma(\lambda_{ij} + N_{ij})} \prod_{k \in V_i} \frac{\Gamma(\lambda_{ijk} + N_{ijk})}{\Gamma(\lambda_{ijk})} \quad (1)$$

Here, the variable $i$ ranges over all of the $n$ nodes of the graph, $j$ ranges over all configurations of the parents of $x_i$, and $k$ over all possible values of $x_i$, which I denote the set $V_i$. The set $C(x_i)$ which $j$ ranges over is defined as a Cartesian product, $C(x_i) = \prod_{\text{Pa}(x_i)} V_i$. $N_{ijk}$ is the count of all data instances where $x_i = k$ and the parents of $x_i$ are in state $j$, while $N_{ij} = \sum_k N_{ijk}$. Similarly, $\lambda_{ijk}$ is a hyperparameter called a pseudocount, the set of which defines the effect of our prior on the score when the network parameters (the CPTs) are integrated out, and $\lambda_{ij} = \sum_k \lambda_{ijk}$.

### 2.3. Gaussian process regression

In previous work (Yackley et al., 2008), we showed that a spline-based regression model could be used to estimate the BDe score of a network. However, this particular model turned out to be unsuitable for search; while the values it returned were very close to the exact ones, the gradients (i.e. the differences between the scores of graphs differing in one edge) were mostly wrong. Motivated by this failure, we tried a different approach. Gaussian process regression is both mathematically simpler than the previous model and gets the gradients mostly correct.

The form of Gaussian process regressor (Rasmussen, 2004) we use is known as simple kriging, and takes the form:

$$\hat{y} = K(g, \hat{g}) K(g, g)^{-1} y \quad (2)$$

In this equation, $g \subseteq \mathbb{X} = \{G_1, G_2, \ldots, G_{n_s}\}$ is a set of $n_s$ training objects (graphs defining Bayesian networks, in our case) with all of the $y$s being their corresponding real-valued scores. $K$ is a function that produces a kernel matrix such that $[K(g, h)]_{ij} = k(g_i, h_j)$, where the positive-definite kernel function $k : \mathbb{X} \times \mathbb{X} \to \mathbb{R}$ maps pairs of objects to a value which can be seen as a generalized inner product; the more alike the two objects are, the higher this value will be. $\hat{g}$ is the new graph (or set of graphs) we are trying to approximate a score for, and $\hat{y}$ is that resulting score. Once the training data is scored, the matrix $K(g, g)^{-1}$ need only be calculated once; from then on, finding an approximate score for any previously-unseen graph is just a matter of calculating $K(g, \hat{g})$ and performing the matrix multiplications.

Finding a proper kernel function on a given set $\mathbb{X}$ is, in general, not a trivial task. However, because our objects are graphs which will always be on the same ordered collection of nodes, we can compare each of the $\binom{n}{2}$ possible edges directly between the two graphs. The form our kernel function takes is:

$$k(G_1, G_2) = \sum_e w_e I[e \in G_1 \wedge e \in G_2] \quad (3)$$

The sum runs over all possible edges of the graph, adding a weight $w_e$ to the kernel's value if that edge is present in both graphs. The weights are tuned using the marginal likelihood gradient [1]; although this process involves repeatedly taking a matrix inverse until the values converge, the size of this matrix is only $n_s \times n_s$, and thus, with a small number of training samples, this is relatively fast.

## 3. Motivation

To do fast structure learning, we want to create a proxy to the exact score function, and this proxy must have two key traits - it must be quick to evaluate, and it must be a good approximation to the true function. Using a Gaussian process regressor gets us the first; once trained, its calculation is a simple matrix product. To get the second, however, we need to know that the true function we are approximating is smooth enough for a Gaussian process to model. This requires, in turn, that we define some topology over the set of directed graphs over which we can say the function is smooth.

---

[1] See (Rasmussen, 2004), Equation 5.9

The topology we use here, which we call the metagraph (Yackley et al., 2008), is defined as the graph of some relation over a set of combinatorial objects. In this case, the objects are themselves directed graphs, and the relation between them is that of differing in exactly one edge. It has two desired properties that make it attractive as a topology over which to search. First, the edges correspond to the search operations we perform - addition and deletion of edges of the target graph. Second, the structure is highly symmetric, taking the form of a hypercube with dimension equal to the possible number of edges of the target. Note that, although they are not valid as Bayesian networks, the metagraph nevertheless includes graphs which contain loops. This is not a problem for an approximator; none of the training structures will contain loops, and a search will still be constrained to that part of the space with no loops.

Furthermore, there is no danger of the approximator being asked to score a graph with cycles (even though the approximation would work mathematically, the answer it returned would be meaningless). Between any two acyclic graphs, a path must exist which never encounters a graph with a cycle; this is trivially proven by considering the process of removing every edge from the first target graph, resulting in a graph with no edges, and then adding back all edges in the second. In general, a shorter path will exist, but this serves as a proof that the region of the metagraph corresponding only to acyclic graphs is fully connected.

## 4. Analysis of smoothness of BDe score

### 4.1. Notation

Let the data set $D \in \mathbb{N}^{m \times n}$ denote a data matrix of discrete values consisting of $m$ i.i.d. observations of $n$ variables. Denote a Bayesian network over these variables as having the graph $G$ and parameters $\Theta$, where $G = <X, E>$ and $X$, the set of variables equals $\{x_1, x_2, \ldots x_n\}$. A score function $sc(G|D)$ maps graphs onto real numbers given a fixed data set, with the convention that a higher score denotes a graph modeling a better explanation of the data. Each variable $x_i$ has a corresponding finite set of possible values $V_i$ and a possibly-empty set of parents in the graph $\text{Pa}(x_i)$. The set of parent configurations $C_i$ for node $x_i$ is given by the Cartesian product $C_i = \prod_{x_j \in \text{Pa}(x_i)} V_j$. The notation $N_{ijk}$ denotes the count across the entire data set of the number of instances where $x_i = k$ and each variable in $\text{Pa}(x_i)$ takes on a value as given by configuration $j \in C_i$. Also, $N_{ij} = \sum_{k \in V_i} N_{ijk}$.

The hyperparameter $\lambda_{ijk}$, needed for the BDe function below, indicates the strength of prior beliefs on the score of a network, needed for a proper Bayesian formulation. As with the $N$s, $\lambda_{ij} = \sum_{k \in V_i} \lambda_{ijk}$.

### 4.2. Basic Definitions

Consider the standard definition of the BDe score, as given in equation 1. In practice, we are more concerned with its logarithm:

$$\log sc(G|D) = \sum_i \sum_{j \in C_i} \Bigg( \log \Gamma(\lambda_{ij}) - \log \Gamma(\lambda_{ij} + N_{ij}) \\ + \sum_{k \in V_i} (\log \Gamma(\lambda_{ijk} + N_{ijk}) - \log \Gamma(\lambda_{ijk})) \Bigg) \tag{4}$$

We assume here that the form of the prior is such that $\lambda_{ijk}$ is equal for all $k$ given a fixed $i$ and $j$, and that this value is inversely proportional to the cardinality of $C_i$. In other words, $\lambda_{ij} = \sum_k \lambda_{ijk} = \#(V_i)\lambda_{ijk}$. Assuming that all nodes are binary, then we simply have $\lambda_{ij} = 2\lambda_{ijk}$ for all $i$ and $j$, and all subscripted $\lambda$s are proportional to some base $\lambda$. Note also that if all nodes are binary, then $\#(V_i) = 2$ for all $i$, and $\#(C_i) = 2^{\#(\text{Pa}(x_i))}$.

In order to prove smoothness, we wish to find upper and lower bounds on the magnitude of the change in score given the addition or deletion of an edge in the graph. Without loss of generality, assume that we add an edge. Call the graph before addition $G$, and the graph after addition $G'$, with scores $sc$ and $sc'$ given the same data set $D$. Because the score takes the form of a sum over all nodes of the graph, the difference between $sc$ and $sc'$ can be captured solely by a single term of the outermost sum, representing the node the new arc points to – call this node $x_\Delta$. We can therefore drop the $i$ subscripts in the formula itself, and represent the one differing term of the two sums using $sc_\Delta$ and $sc'_\Delta$ respectively. Because all other terms of the sum remain unchanged, $sc - sc' = sc_\Delta - sc'_\Delta$. Also, the range of the initial $j$ variable, $C_\Delta$, now splits into two sets, $C_0$ and $C_1$, where the subscript indicates the value of the newly added parent. For each element of $C_\Delta$, there is a corresponding element both in $C_0$ and $C_1$, and $\#(C_\Delta) = \#(C_0) = \#(C_1)$.

### 4.3. Form of the bound on $sc_\Delta - sc'_\Delta$

From the above, we have:
$$sc_\Delta = \sum_{j \in C_\Delta} \Bigg( \log \Gamma(\lambda_j) - \log \Gamma(\lambda_j + N_j) +$$
$$\sum_{k \in V} (\log \Gamma(\lambda_{jk} + N_{jk}) - \log \Gamma(\lambda_{jk})) \Bigg)$$
$$= \sum_{j \in C_\Delta} \big( \log \Gamma(\lambda_j) - \log \Gamma(\lambda_j + N_j) + \log \Gamma(\lambda_{j0} + N_{j0})$$
$$- \log \Gamma(\lambda_{j0}) + \log \Gamma(\lambda_{j1} + N_{j1}) - \log \Gamma(\lambda_{j1}) \big)$$

Since $\lambda_{j0} = \lambda_{j1} = \lambda_j / 2$ and $N_j = N_{j0} + N_{j1}$, we can simplify this to:
$$sc_\Delta = \sum_{j \in C_\Delta} (\alpha - \log \Gamma(\lambda_j + N_{j0} + N_{j1}) + \log \Gamma(\frac{\lambda_j}{2} + N_{j0}) + \log \Gamma(\frac{\lambda_j}{2} + N_{j1})), \quad (5)$$

where $\alpha = \log \Gamma(\lambda_{ij}) - 2 \log \Gamma(\lambda_{ij}/2)$. The only difference between $sc_\Delta$ and $sc'_\Delta$ is the set over which $j$ ranges; if we abbreviate the preceding sum as $sc_\Delta = \sum_{j \in C} f(j)$, then $sc'_\Delta = \sum_{j \in C_0 \cup C_1} f(j) = \sum_{j \in C_0} f(j) + \sum_{j \in C_1} f(j)$. Note, however, that the two addends each take the same form as the expression for $sc_\Delta$, and that the sets $C_0$ and $C_1$ have the same size as $C$, with all elements in a one-to-one correspondence. With some relabeling of variables, we have:
$$sc_\Delta - sc'_\Delta = \sum_{j \in C} f(j) - \sum_{j_0 \in C_0} f(j_0) - \sum_{j_1 \in C_1} f(j_1)$$
$$= \sum_{j \in C_\Delta} (f(j) - f(j_0) - f(j_1))$$

Some notation abuse takes place in the second equation; $j_0$ and $j_1$ are the corresponding configurations in $C_0$ and $C_1$ to $j$ in $C_\Delta$, with the value in the additional parent being 0 or 1 respectively. Because of this, $N_{j_0 k} + N_{j_1 k} = N_{jk}$ for any $k$, and likewise $N_{j_0} + N_{j_1} = N_j$. Also, we have $\lambda_{j_0} = \lambda_{j_1} = \lambda_j / 2$. Corresponding to the above definition of $\alpha$, let $\beta = \log \Gamma(\lambda_{ij}/2) - 2 \log \Gamma(\lambda_{ij}/4)$. Even making these simplifications, the full expression for $sc_\Delta - sc'_\Delta$ expands to a cumbersome form; to simplify it further, we introduce an auxiliary function denoted as $\gamma$.

### 4.4. The function $\gamma(a, b)$

Let the function $\gamma(a, b)$ be defined as follows[2]:
$$\gamma(a, b) = \log \Gamma(a+b) - \log \Gamma(a) - \log \Gamma(b) \quad (6)$$

---
[2]This function is related to the standard Beta function; $\gamma(a, b) = -\log \mathrm{B}(a, b)$

Using Stirling's approximation for the log-gamma function (Abramowitz & Stegun, 1964) ($\ln x! = x \ln x + x - \Theta(x)$), we obtain a result which will be important later:
$$\gamma(a, a) = 2a \log(2a) - 2a + \Theta(\log 2a)$$
$$- 2(a \log a - a + \Theta(\log a))$$
$$= (2 \log 2)a + \Theta(\log a) \quad (7)$$

Now, we can use $\gamma$ to simplify the equation for $sc_\Delta - sc'_\Delta$, given that $N_{j_0 0} + N_{j_1 0} = N_{j0}$ and $N_{j_0 1} + N_{j_1 1} = N_{j1}$. We also split out the term inside the sum and call it $t$, for reasons given below.
$$t = \gamma(\frac{\lambda_j}{4} + N_{j_0 0}, \frac{\lambda_j}{4} + N_{j_1 0})$$
$$+ \gamma(\frac{\lambda_j}{4} + N_{j_0 1}, \frac{\lambda_j}{4} + N_{j_1 1}) \quad (8)$$
$$- \gamma(\frac{\lambda_j}{2} + N_{j_0 0} + N_{j_0 1}, \frac{\lambda_j}{2} + N_{j_1 0} + N_{j_1 1})$$
$$sc_\Delta - sc'_\Delta = 2^{\#(Pa(x_i))}(\alpha - 2\beta) + \sum_{j \in C_\Delta} t \quad (9)$$

### 4.5. Getting to the extrema

We seek upper and lower bounds on $sc_\Delta - sc'_\Delta$ given fixed $\lambda_j$ (and therefore fixed $\alpha$ and $\beta$ as well). We therefore differentiate the equation with respect to the four $N$s and set the four derivatives all equal to zero. Because the sum over $j \in C_i$ is irrelevant (a sum over any number of worst cases will produce a worst case, and likewise for best cases), we only need to calculate bounds for $t$, which we will accomplish by taking its derivative with respect to the four $N_{pq}$ variables to find its minimum and maximum.

Because $t$ is defined in terms of $\gamma$, which is itself defined in terms of the $\log \Gamma$ function, the results will involve the $\psi$ function[3]. For space reasons, we abbreviate expressions of the form $\psi(\frac{\lambda_j}{2} + N_{j_a b} + N_{j_c d})$ as $\psi_{ab,cd}$. $\psi_{ab}$ stands for $\psi(\frac{\lambda_j}{4} + N_{j_a b})$, and $\psi$ alone stands for $\psi(\lambda_j + N_{j_0 0} + N_{j_0 1} + N_{j_1 0} + N_{j_1 1})$. Using these abbreviations, the derivative of $t$ with respect to some $N_{ab}$ is:

$$\frac{dt}{dN_{j_a b}} = -\psi_{ab} + \psi_{a0,a1} + \psi_{0b,1b} - \psi \quad (10)$$

Taking the four derivatives of $t$ and setting them equal to zero, we obtain the system of equations:
$$\psi_{00} + \psi = \psi_{00,01} + \psi_{00,10}$$
$$\psi_{01} + \psi = \psi_{00,01} + \psi_{01,11}$$
$$\psi_{10} + \psi = \psi_{10,11} + \psi_{00,10}$$
$$\psi_{11} + \psi = \psi_{10,11} + \psi_{01,11}$$
$$\quad (11)$$

---
[3]Defined the standard way as $\psi(x) = \frac{d}{dx} \log \Gamma(x)$

By subtracting pairs of equations, we obtain:
$$\begin{aligned}
\psi_{00} - \psi_{01} &= \psi_{00,10} - \psi_{01,11} \\
\psi_{00} - \psi_{10} &= \psi_{00,01} - \psi_{10,11} \\
\psi_{01} - \psi_{11} &= \psi_{00,01} - \psi_{10,11} \\
\psi_{10} - \psi_{11} &= \psi_{00,10} - \psi_{01,11}
\end{aligned} \quad (12)$$

One solution is apparent from inspection. If we set $N_{j_00} = N_{j_10} = N_{j_01} = N_{j_11}$, then all four equations reduce to $0 = 0$. One of our extrema, therefore, occurs there, corresponding to the case where we add an edge to split apart data which is already uniformly distributed in both variables corresponding to the edge's endpoints. In other words, this edge has no reason to exist in a Bayesian network, and should logically decrease the score by the most; this is a maximum.

$$\begin{aligned}
\max sc_\Delta - sc'_\Delta &= 2^{\#(Pa(x_i))}(\alpha - 2\beta) \\
&+ \sum_{j \in C_i} \left( 2\gamma(\frac{\lambda_j}{4} + N_{j_00}, \frac{\lambda_j}{4} + N_{j_00}) \right. \\
&\left. - \gamma(\frac{\lambda_j}{2} + 2N_{j_00}, \frac{\lambda_j}{2} + 2N_{j_00}) \right)
\end{aligned} \quad (13)$$

Since we are only concerned with the asymptotic behavior of this function, we can drop the constant terms as well as the summation (which is over a constant number of terms independent of the value of any of the $N$s).

$$\begin{aligned}
\max sc_\Delta - sc'_\Delta = O\left( 2\gamma(\frac{\lambda_j}{4} + N, \frac{\lambda_j}{4} + N) \right. \\
\left. - \gamma(\frac{\lambda_j}{2} + 2N, \frac{\lambda_j}{2} + 2N) \right)
\end{aligned} \quad (14)$$

From equation 7, we obtain:
$$\begin{aligned}
\max sc_\Delta - sc'_\Delta &= O\left( (4 \log 2)(\lambda_j/4 + N) \right. \\
&\left. - \Theta(\log N) - (2 \log 2)(\lambda_j/2 + 2N) + \Theta(\log N) \right) \\
&= O((\log 2)(\lambda_j + 4N) - (\log 2)(\lambda_j + 4N) + \Theta(\log N)) \\
&= O(\log N)
\end{aligned}$$

This indicates that, in cases where adding an edge lowers the score, the worst it can lower it by is only logarithmic in the number of data points.

The other solutions to the system occur where $N_{j_10} = N_{j_01} = 0$ or $N_{j_00} = N_{j_11} = 0$, representing data which (in our binary-variable case) is perfectly aligned in such a way that both the marginal of the node and its new parent seem uniform, but adding the edge reveals their values to be in perfect correspondence with one another. The reasoning behind is is as follows.

|       | $j=0$ | $j=1$ |
|-------|-------|-------|
| $k=0$ | 0.5   | 0     |
| $k=1$ | 0     | 0.5   |

$$t = \gamma(\frac{\lambda_j}{4} + \frac{N}{2}, \frac{\lambda_j}{4}) + \gamma(\frac{\lambda_j}{4}, \frac{\lambda_j}{4} + \frac{N}{2}) - \gamma(\frac{\lambda_j}{2} + \frac{N}{2}, \frac{\lambda_j}{2} + \frac{N}{2})$$

|       | $j=0$ | $j=1$ |
|-------|-------|-------|
| $k=0$ | 0.25  | 0.25  |
| $k=1$ | 0.25  | 0.25  |

$$t = 2\gamma(\frac{\lambda_j}{4} + \frac{N}{4}, \frac{\lambda_j}{4} + \frac{N}{4}) - \gamma(\frac{\lambda_j}{2} + \frac{N}{2}, \frac{\lambda_j}{2} + \frac{N}{2})$$

Figure 1. Illustrations of the best and worst cases for $sc_\Delta - sc'_\Delta$, in the form of joint probability tables

Consider our expression for $t$ above. The minimum value occurs when the first two (positive) terms of the sum are minimized and the negative term is maximized. Because we know from section 4.4 that the $\gamma$ function is maximized when the arguments are equal and minimized when they are farthest apart, we can force this to happen by setting $N_{j_10} = N_{j_01} = 0$ or $N_{j_00} = N_{j_11} = 0$ and the other two variables equal to one another. This case corresponds to having a marginal distribution over both variables which is uniform, but where the joint indicates a perfect correspondence between the two. This is exactly the sort of situation where an edge ought to be added.

$$\begin{aligned}
\min sc_\Delta - sc'_\Delta = O\left( \gamma(\frac{\lambda_j}{4} + N, \frac{\lambda_j}{4}) \right. \\
\left. + \gamma(\frac{\lambda_j}{4}, \frac{\lambda_j}{4} + N) - \gamma(\frac{\lambda_j}{2} + N, \frac{\lambda_j}{2} + N) \right)
\end{aligned}$$

Because $\gamma(a, b)$ is maximized for a fixed $a + b$ when $a = b$, we can say that $\gamma(\lambda_j/4 + N, \lambda_j/4) < \gamma(\lambda_j/4 + N/2, \lambda_j/4 + N/2)$, and so

$$\begin{aligned}
\min sc_\Delta - sc'_\Delta &< O\left( 2\gamma(\frac{\lambda_j}{4} + \frac{N}{2}, \frac{\lambda_j}{4} + \frac{N}{2}) \right. \\
&\left. - \gamma(\frac{\lambda_j}{2} + N, \frac{\lambda_j}{2} + N) \right) \\
&= O((4 \log 2)(\lambda_j/4 + N/2) \\
&\quad - (2 \log 2)(\lambda_j/2 + N) + \Theta(\log N)) \\
&= O((\log 2)(\lambda_j + 2N) \\
&\quad - (\log 2)(\lambda_j + 2N) + \Theta(\log N)) \\
&= O(\log N)
\end{aligned}$$

Both the minimum and maximum score jumps, then, are simply logarithmic in the number of data points, showing that, with respect to a topology derived from addition and deletion of edges, the BDe score is Lipschitz smooth with a constant of $K = O(\log N)$.

### 4.6. Implications

As one would expect, the worst case scenario is to add an edge that provides no information at all. If the joint distribution between $x_i$ and its new parent is uniform, the model gains nothing by putting the edge there, while the score (as it should) penalizes the addition. The best case, meanwhile, is for the new edge to link $x_i$ to a parent that perfectly matches its values (or at least a permutation of them) in all cases, while the marginals of the joint distribution are entirely uniform and uninformative. These fit our intuitions of how edges in a Bayesian network should be interpreted. Also, because the worst possible changes to the score are merely logarithmic in the size of the data set, the search landscape is sufficiently smooth that a Gaussian Process regressor is an appropriate choice to represent it.

The Gaussian process regressor is a good choice for another reason – the fact that it is based on a kernel function means that its complexity is not based on the size of the training set or the size of the graphs (or, for that matter, the size of the original data set), but the VC dimension of the kernel space (Schölkopf & Smola, 2001).

Note also that, once the training set is scored, there is no longer a need to keep around the original data set – all of the information we need to search has been encapsulated into the proxy. This is a clear win in the case where the data set has a large number of instances; instead of needing to count up values for $N_{ijk}$ across perhaps millions of data points every time we take a search step, we can simply refer to the proxy.

### 4.7. Other score functions

It is an open question, and one we hope to address in the future, whether the same kind of smoothness bound can be proven for other Bayesian network score functions. For example, the BIC score (Schwarz, 1978) is defined as follows, in terms of a log-likelihood score and a penalty term.

$$sc_{BIC}(G|D) = \sum_{i=1}^{n} \sum_{j \in C_i} \sum_{k \in V_i} N_{ijk} \log\left(\frac{N_{ijk}}{N_{ij}}\right) - \frac{1}{2}\log(m)|B| \quad (15)$$

$|B| = \sum_{i=1}^{n}(\#V_i - 1)\#C_i$ is the number of degrees of freedom across the parameter set $\Theta$. In this form, adding an edge to a network will split the set of parent configurations, as before, by adding another term to the product which defines $C_i$. However, it will also alter the value of the penalty term $|B|$.

## 5. Proxy-Accelerated Search Results

To compare the effects of the proxy to an exact-scoring search, we selected six data sets on which to build Bayesian networks. Three of these, ADULT1, ADULT2, and ADULT3, came from the original paper that introduced the ADTree (Anderson & Moore, 1998), where they were used as examples of data sets an ADTree could be built on. The ADTree is a structure which provides a caching mechanism to accelerate the process of scoring; it trades off an initial tree-build time and the memory needed to store the structure to achieve much faster speed at the kind of $N_{ijk}$ counts necessary to compute a BDe score. The results for those three data sets show that, even with ADTree-based acceleration, we are able to find comparable scores in much less time using the proxy. The proxy-based search was performed 5 times with randomly selected training samples each time; the results shown here are the mean and standard deviation. The algorithm was a standard greedy search, chosen to be a reasonable baseline. It should be mentioned, though, that the benefit of using a proxy would extend, in theory, to any search algorithm that uses a scoring function.

The scores of the graphs as reported in the table are exact, not derived from the proxy. Although the values that the proxy returns are often very far off from exact, the gradients remain intact, and this is why we can count on the proxy to drive a search in the right direction. The values for $n_s$ reported in Table 5 are those for which our proxy performed best; experiments were conducted for a small range of different values for $n_s$.

The other three data sets are taken from the UCI Data Repository (Frank & Asuncion, 2010); they are CENSUS-INCOME, TIC2000, and MUSK. All of these are too large for an ADTree to fit in memory, and so the scores were calculated using the Bayes Net Toolkit (Murphy, 2001) and its accompanying Structure Learning Package (Leray & Francois, 2004). The proxy-based searches on CENSUS-INCOME and TIC2000 were performed five times, as above, while the MUSK data set was large enough that it was only practical to perform a single search for each differing number of training samples. The algorithms were implemented in MATLAB, on a Linux server running at 2.2 GHz with 32 gigabytes of RAM.

### 5.1. Discussion

The effects of the proxy are clear; in all but one case, the networks found by the proxy-based search were either comparable to or significantly better than those found by the exact-scoring version, and always in a shorter time. At present, we don't know what property of the CENSUS-INCOME data set made it perform so poorly.

|         | $n$ | $m$   | $n_s$ |
|---------|-----|-------|-------|
| ADULT1  | 15  | 15060 | 250   |
| ADULT2  | 15  | 30162 | 250   |
| ADULT3  | 15  | 45222 | 100   |
| CENS-INC| 42  | 95130 | 250   |
| TIC2000 | 86  | 9822  | 25    |
| MUSK    | 168 | 6598  | 60    |

Table 1. Data set properties and numbers of samples used to train the approximator in each case.

|         | time (std.) | time (GPR)       |
|---------|-------------|------------------|
| ADULT1  | 91.32       | $22.70 \pm 1.36$ |
| ADULT2  | 149.75      | $34.90 \pm 2.27$ |
| ADULT3  | 209.59      | $18.00 \pm 0.20$ |
| CENS-INC| 733.4       | $501.4 \pm 41.3$ |
| TIC2000 | 45.2        | $23.24 \pm 0.83$ |
| MUSK    | 3907        | 666.5            |

Table 2. Time summary. All times are in seconds.

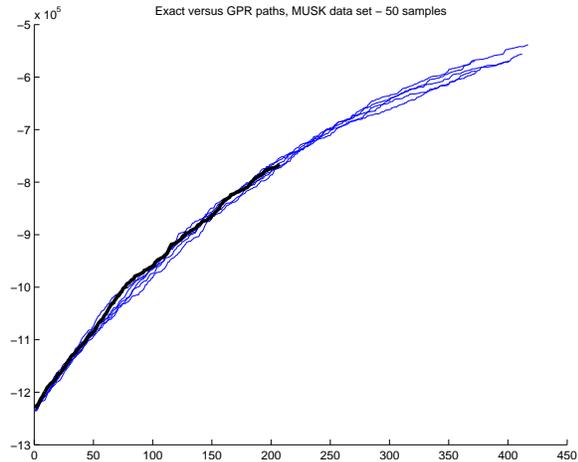

Figure 2. Trajectories for $n_s = 50$, MUSK data set

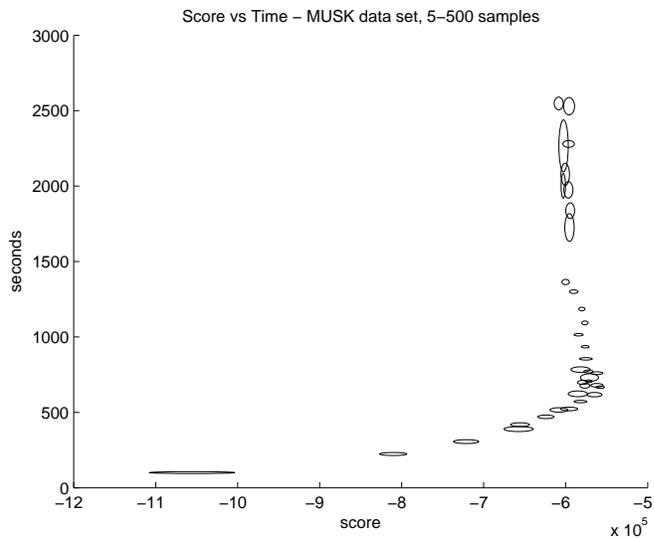

Figure 3. Time as a function of score, MUSK data set

In every other case, however, the advantage of the smoothing induced by the proxy is clear, and this is most dramatic in the case of the MUSK data set. With a relatively tiny number samples across the immense space of networks on 168 nodes, the proxy was nevertheless able to find a network with a greatly improved score. The reason for this — and the reason smoothness is so important — is shown in Figure 5.1. These lines are the search trajectories, with search step on the $x$ axis and score on the $y$ axis. The thick line is the trajectory taken by the exact-scoring search, while the thinner blue lines are the ones taken by five runs of the proxy with different sets of 50 training samples. The exact search stops partway through, having encountered a local maximum. However, the proxy will tend to smooth these local features out, letting the search process continue to greater heights. In fact, too many training samples can in fact hamper the proxy's even-tual score. Figure 5.1 plots each of the proxy's runs with a different value of $n_s$ as a bubble in a time-score graph, with the size of the oval being one standard deviation in either dimension. The lowest value, $n_s = 5$, is at the bottom left, having taken a very short overall time but producing a relatively bad network. The bubbles continue up and to the right, with both time and score growing, until we reach the farthest-right point on the graph when $n_s = 60$. From there, the time continues to increase, but the score worsens. We believe that this is due to the proxy starting to learn the space too well, capturing the finer features of the score landscape while losing sight of the bigger picture.

|         | score (std.) | score (GPR)        |
|---------|--------------|--------------------|
| ADULT1  | $-1.219$     | $-1.265 \pm 0.028$ |
| ADULT2  | $-3.059$     | $-2.964 \pm 0.104$ |
| ADULT3  | $-4.879$     | $-4.679 \pm 0.169$ |
| CENS-INC| $-8.205$     | $-9.614 \pm 0.66$  |
| TIC2000 | $-3.498$     | $-3.050 \pm 0.09$  |
| MUSK    | $-7.659$     | $-5.579$           |

Table 3. Score summary. Scores are $\times 10^5$.

## 6. Future Work

We are currently working on extending the proxy to other score-based search strategies, such as simulated annealing (Kirkpatrick et al., 1983), as well as to other combinatorial objects such as general 0-1 matrices and permutations. The success of these rests, it would seem, on finding a proper form for a kernel function on these objects, thus defining the topology of the space both traversed by the search method and used by the approximator.

Another direction we wish to extend this in is to implement the training phase on a massively parallel system, which would greatly reduce the time taken to train the proxy. This would also require the implementation of a way to combine the training results; a block-matrix inversion technique will be useful here, as well as adding the potential to add more training data in the middle of an ongoing search. This way, the space around an apparent local maximum could be examined in greater detail and refined.

## 7. Conclusion

As data sets increase in size, it becomes more necessary to develop algorithms which can search for and identify models of them in reasonable amounts of time. However, the larger the data set gets, the more time this takes, and the larger the search space, the more chance there is of a search running into a local maximum instead of the desired global. A proxy function will alleviate both of these problems; in particular, we showed that the BDe score considered over a search space of single-edge additions and deletions is smooth enough to make a proxy-based search viable, and the results bear this out.

This process, building a proxy function from a set of random samples and then using it to drive a search, is readily applicable to any search algorithm that depends on calculating a series of scores, from a simple greedy search to more sophisticated ones such as Markov Chain Monte Carlo. These new accelerated forms of algorithms will allow researchers in fields as diverse as astronomy (Kent, 1994), biology (Roy et al., 2007), and linguistics (Davies, 2009) to better analyze data and create hypotheses given their often staggeringly large data sets. Through the use of the proxy-based search accelerator, we will be able to find patterns in more complex data than had previously been feasible.

## 8. Acknowledgements

The authors wish to thank Blake Anderson and Eduardo Corona for their ideas and support, as well as the Machine Learning Reading Group at the University of New Mexico. This research was supported by National Science Foundation grant IIS-0705681 and Office of Naval Research grant N000141110139.